\documentclass[letterpaper, 10 pt, conference]{ieeeconf}  % Comment this line out if you need a4paper

\IEEEoverridecommandlockouts                              % This command is only needed if 
\usepackage{hyperref}
\usepackage{caption}
\usepackage{subcaption}
\usepackage{xcolor}
\usepackage[nospace]{cite}
\usepackage{booktabs}
\usepackage{multirow}
\usepackage{tikz}
\usetikzlibrary{positioning}
\usepackage{comment}
\usepackage{amsmath}
\DeclareMathOperator*{\argmax}{arg\,max}
\DeclareMathOperator*{\argmin}{arg\,min}
\usepackage{amsmath,bm}

\usepackage{upgreek}
\title{\LARGE \bf
Active Visual Localization for Multi-Agent Collaboration:\\A Data-Driven Approach
}

\author{Matthew Hanlon*\quad Boyang Sun*\quad  Marc Pollefeys\quad Hermann Blum% <-this % stops a space
\thanks{*equal contribution}% <-this % stops a space
\thanks{All authors are with the Computer Vision and Geometry Lab at ETH Zürich.}%
}

\begin{document}

\maketitle
\thispagestyle{empty}
\pagestyle{empty}

%%%%%%%%%%%%%%%%%%%%%%%%%%%%%%%%%%%%%%%%%%%%%%%%%%%%%%%%%%%%%%%%%%%%%%%%%%%%%%%%
\begin{abstract}
Rather than having each newly deployed robot create its own map of its surroundings, the growing availability of SLAM-enabled devices provides the option of simply localizing in a map of another robot or device. In cases such as multi-robot or human-robot collaboration, localizing all agents in the same map is even necessary. However, localizing e.g. a ground robot in the map of a drone or head-mounted MR headset presents unique challenges due to viewpoint changes. This work investigates how active visual localization can be used to overcome such challenges of viewpoint changes. Specifically, we focus on the problem of selecting the optimal viewpoint at a given location. We compare existing approaches in the literature with additional proposed baselines and propose a novel data-driven approach. The result demonstrates the superior performance of our data-driven approach when compared to existing methods, both in controlled simulation experiments and real-world deployment.
\end{abstract}

%%%%%%%%%%%%%%%%%%%%%%%%%%%%%%%%%%%%%%%%%%%%%%%%%%%%%%%%%%%%%%%%%%%%%%%%%%%%%%%%
\section{Introduction}
\label{sec:introduction}

\begin{figure}
\centering
  \centering
  \includegraphics[width=0.45\linewidth]{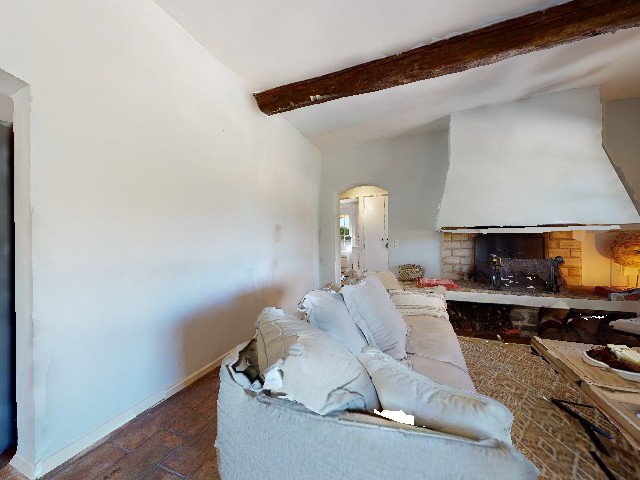}
  \includegraphics[width=0.45\linewidth]{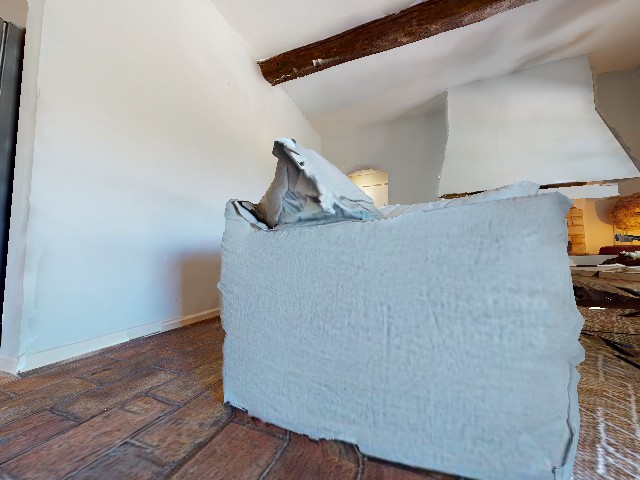}
\caption{\textbf{Difference in perspective} between a head-mounted sensor rig used for mapping (left) and a ground robot (right) deployed for localization.}
\label{fig:perspective_diff}
\vspace{-3mm}
\end{figure}

Visual localization and mapping systems are by now ubiquitous around humans. They are part of every smartphone, to e.g. improve localization in GPS denied environments, they are used in VR and AR headsets, in cars, and of course in robots. In parallel to this rollout, more and more environments get mapped, and localization becomes an interesting cloud service where any agent can send its observation and gets it localized in a pre-existing map. This can even be achieved without violating privacy~\cite{speciale2019privacy}. However, it raises the question how well devices and robots can be localized if the map was created from a different kind of device and possibly from quite different points of view, as illustrated in Figure~\ref{fig:perspective_diff}.

As an example of this larger question, this paper delves into the specific scenario of localizing a new agent, such as a ground robot, in a pre-existing map. This application holds considerable practical value, as it obviates the need to re-map an entire building if a suitable map already exists. Furthermore, for seamless collaboration between human-robot teams, the ability to localize mobile devices and robots within a shared map becomes imperative, as highlighted in previous studies~\cite{sutdhololens, erat2018drone, reardon2018come}. However, accurately localizing a robot in a map created with a head-mounted camera rig introduces its own challenges.
These challenges stem from the use of diverse sensor devices and are compounded by significant variations in viewpoint between the mapping trajectory and the operational height of the robot. Such variations result in multiple causes that diminishes the localization performances, such as reduced visual overlap. This issue becomes particularly prominent in the case of ground robots, including quadrupeds, where obstacles such as chairs, tables, and furniture frequently obscure substantial portions of the robot's environment.

\begin{figure}

\begin{tikzpicture}
\footnotesize
    \node[inner sep=0] (main) {\includegraphics[width=1.0\linewidth]{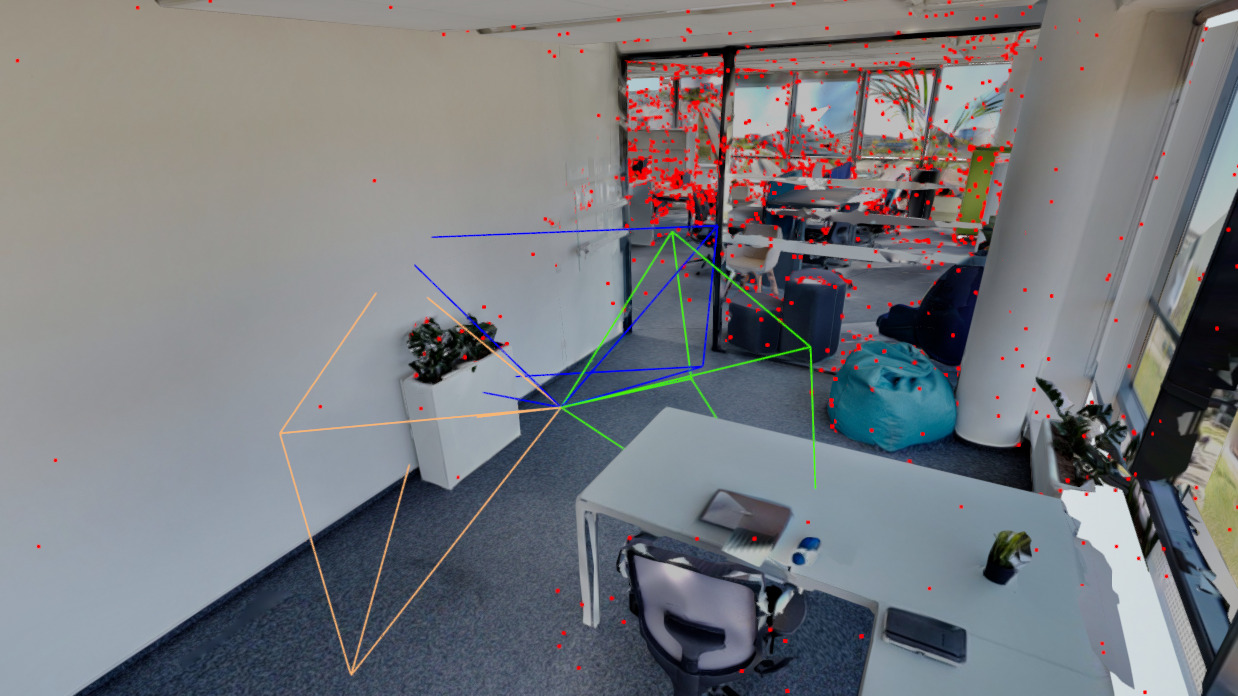}};
    \node[inner sep=0,above=7mm of main.north west, anchor=south west, label={[align=center, text=orange]below:looking forward \\ 3.01m / 106deg}] (forward) {\includegraphics[width=.31\linewidth]{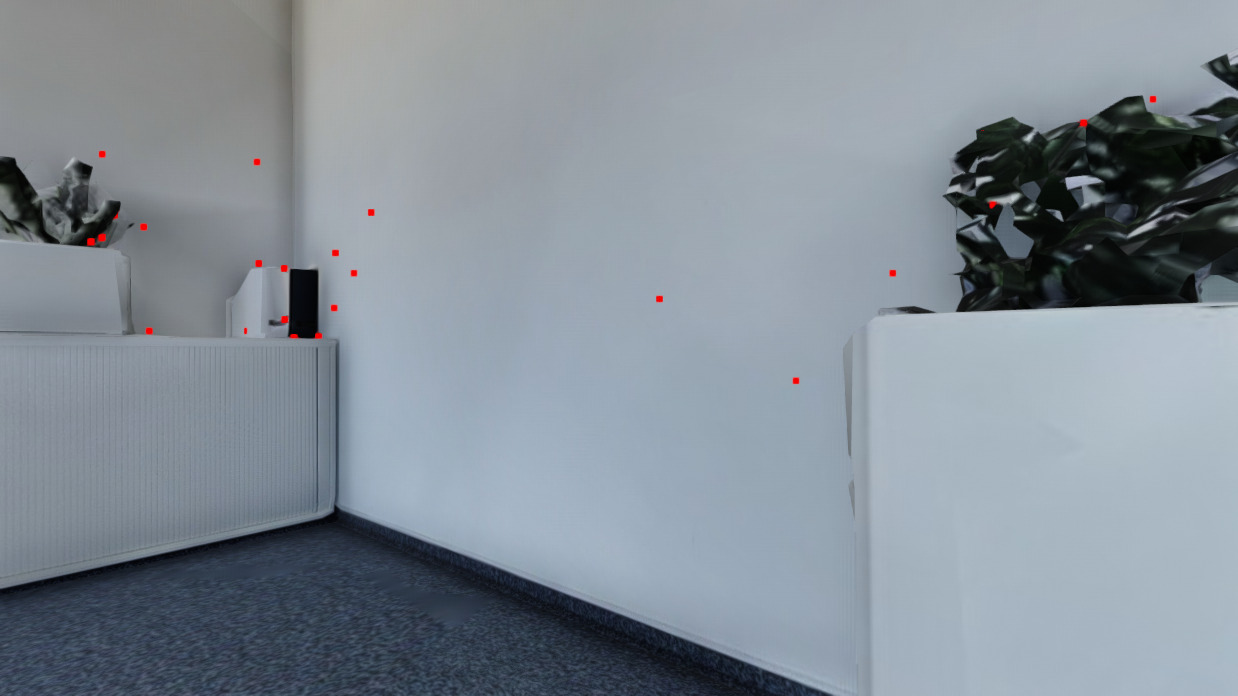}};
    \node[inner sep=0,above=7mm of main,label={[align=center, text=blue]below:viewing angle strategy \\ 0.15m / 0.8deg}] (angle) {\includegraphics[width=.31\linewidth]{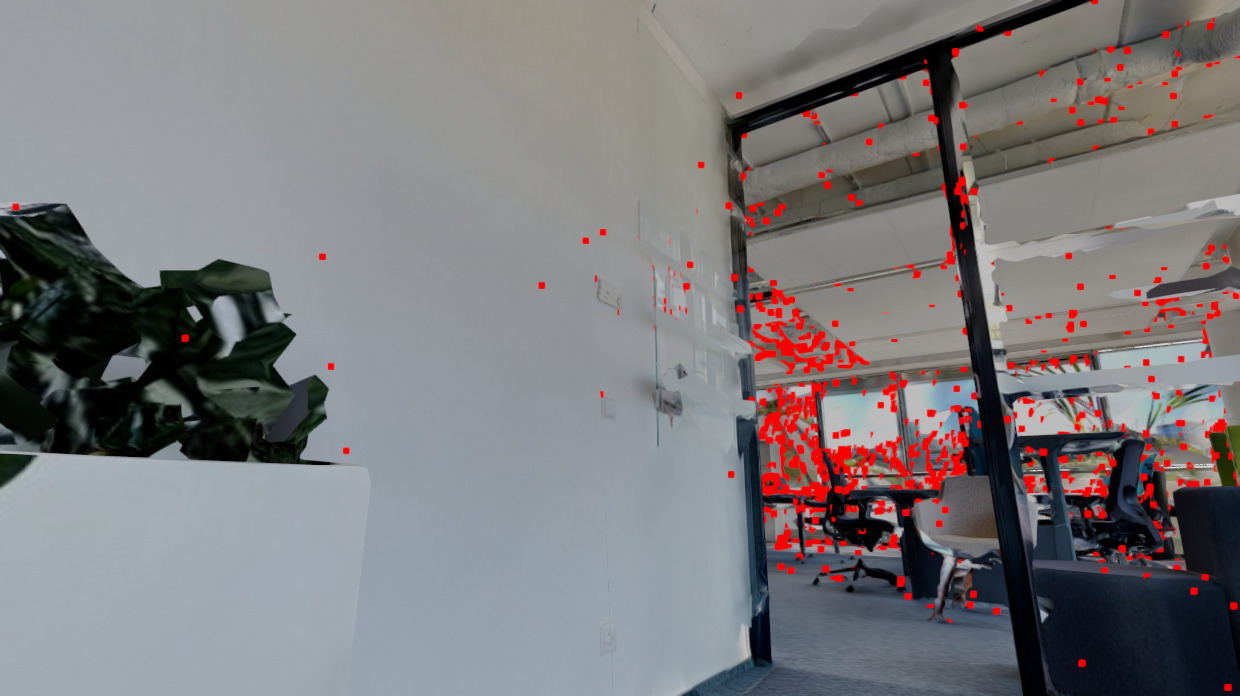}};
    \node[inner sep=0,above=7mm of main.north east, anchor=south east,label={[align=center, text=teal]below:data-driven (ours) \\ 0.01m / 0.1deg}] (transformer) {\includegraphics[width=.31\linewidth]{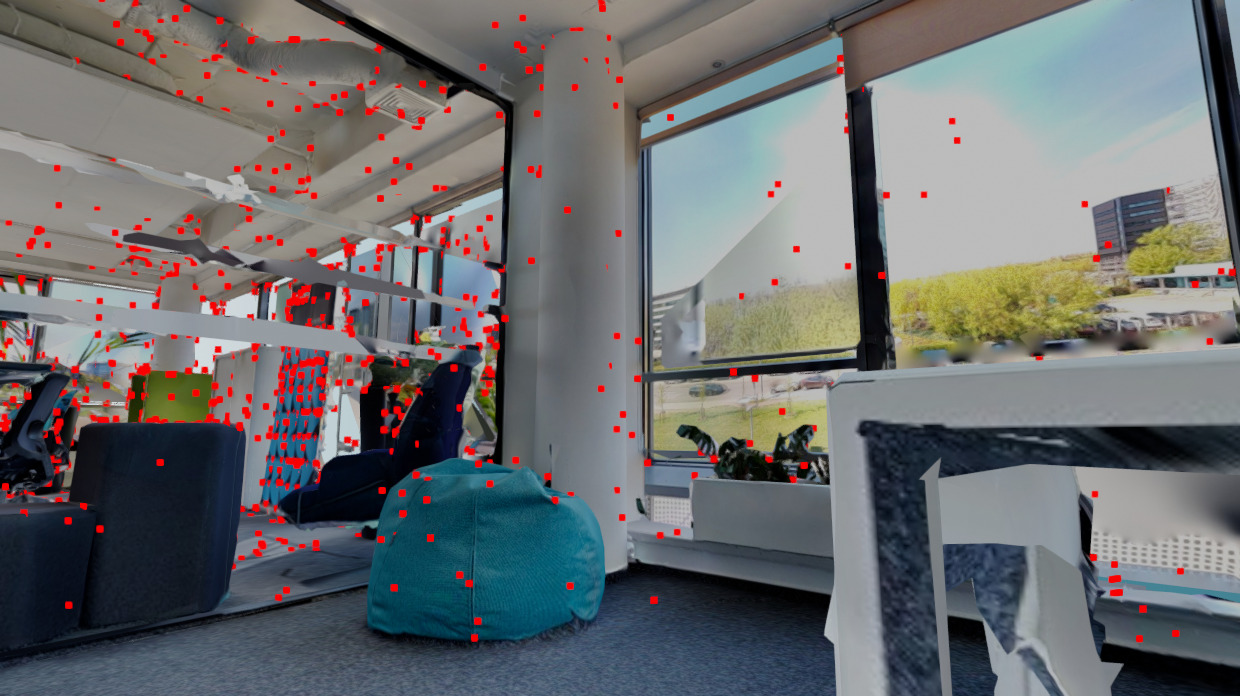}};
\end{tikzpicture}
\caption{\textbf{Viewpoint Selection} of three methods that run visual localization at the same location with respect to the built map (landmarks in red). The passive strategy of looking \textcolor{orange}{\textit{forward}}. and a strategy inspired by~\cite{Davison2002SimultaneousVision} to maximize the similarity with the \textcolor{blue}{\textit{viewing angle}} towards the landmark during mapping both result in higher localization error than our data-driven \textcolor{teal}{\textit{viewpoint transformer (VPT)}} approach.}
\label{fig:qualitative_results}
\vspace{-3mm}
\end{figure}

Many works have studied how to better localize a given image within the map. However, in contrast to always trying to achieve the best from a given viewpoint, robots possess the valuable ability to autonomously select viewpoints. This leads us to investigate whether active viewpoint selection can effectively address the challenges associated with cross-agent visual localization.
In the literature, this concept is widely recognized as active perception, and within our specific context, it is referred to as Active Visual Localization. The core objective of Active Visual Localization is to determine the camera pose within an existing map representation of the environment, in order to improve localization accuracy. A common approach involves assessing the localization utility of various viewpoints, typically through metrics like the Fisher Information Metric (FIM), or a combination of hand-crafted heuristics. While extensive work has been directed towards integrating these utility calculations into planning frameworks, a relatively less studied component is how the utility value itself can be better estimated, particularly in scenarios involving the unique challenges discussed earlier.

This work delves into the exploration of an effective utility function to actively enhance visual localization of a robot in a mapped environment.  Our primary focus lies in evaluating established viewpoint selection and assessment criteria, while introducing a novel data-driven approach to viewpoint scoring. The key contributions of this paper are:
\begin{itemize}
    \item \textbf{A novel data-driven approach} to viewpoint scoring resp. selection for active localization
    \item \textbf{Comparison and thorough evaluation} of viewpoint selection methods for the problem of visual localization between heterogeneous agents
    \item \textbf{Real-world validation} of our findings by integrating the viewpoint selection into an active viewpoint planner for a robot with an arm-mounted camera

\end{itemize}

% We further open-source our implementations and collected datasets.

\section{Related Work}\label{sec:related_work}
Active Vision describes the case where an agent has the ability to move visual sensors with the goal of improving the performance of perceptual tasks~\cite{aloimonos1987weiss}. This concept finds application across various domains where visual information is utilized, such as scene exploration~\cite{schmid2020efficient,chaplot2020learning}, data collection~\cite{ruckin2023informative,ye2022multi}, inspection~\cite{xing_23_autonomous}, and active learning~\cite{zurbrugg2022embodied, ruckin2024semi}. The case studied in this work focuses on the task of improving visual localization by choosing the most informative viewpoint at a given position, which can be considered a subset of Active Vision, named Active Visual Localization. Numerous existing works have studied how informative metrics can be incorporated into motion planning frameworks with the goal of optimizing robot motion to maximize localization accuracy~\cite{roy1999coastal,costante2016perception,zhang2018perception}. Within the scope of this research, our focus is on the specific task of augmenting visual localization by selecting the most informative viewpoint at a given position.

Regarding viewpoint selection in the realm of active vision, early approaches primarily rely on handcrafted metrics to gauge the uncertainty or reliability of a given viewpoint in contributing to the system's state~\cite{Sadat2014Feature-richMono-SLAM,1389909,6631022,fontanelli2009visual,8461133,7989531,10161136}. Some methods take a different path by deriving evaluation metrics from the metric map and constructing a global utility map. Authors of~\cite{zhang2020fisher} propose a solution rooted in Fisher information theory~\cite{feder1999adaptive,makarenko2002experiment}. The central task they explore consists of determining the amount of information a viewpoint from a given pose will contribute to the localization process. They develop a novel map representation that enables efficient computation of the Fisher information for 6-DoF visual localization, known as the Fisher Information Field. Similar ideas of using Fisher Information for viewpoint selection have also been explored in recent works~\cite{10161136,kim2015active,abraham2019active}. However, it's important to note that these metrics often rely on heuristics, necessitate the design of specific handcrafted utility functions, and may have limitations in their representation capabilities for diverse and complex scenarios.

Another category of works introduces additional vision tasks to enhance the metric extraction process, notably incorporating semantic information~\cite{tao20233d}. For instance, the work of~\cite{Bartolomei2021Semantic-awareLearning} aims to improve navigation performance by including semantic information, in order to discern perceptually informative areas of the environment. This kind of work enriches the semantic understanding capability, however highly relies on the performance of the semantic module, and usually requires prior knowledge to link certain semantic classes that clearly correlate to visual informativeness.

A separate line of research focus on using data-driven approaches to active visual localization. These works mostly choose to formulate the problem as a reinforcement learning task,  tightly integrating metric determination with robot execution  \cite{chaplot2018active,fang2022towards}. As an example,~\cite{lodel2022look} trains an information-aware policy to find traversable paths as well as reduce the uncertainty of the environment. The learning-based model significantly enhance the robot's comprehension of its surroundings. However, it's worth noting that these models typically demand substantial computational resources for training and may require the simplification of the environment model to prevent over-fitting.

In this work, we try to combine the advantages of both the data-driven approach and the viewpoint scoring scheme by formulating viewpoint selection as a classification problem.

\section{Method}
\label{sec:method}

\begin{figure*}[htbp]
    \centering
    \includegraphics[width=0.75\linewidth]{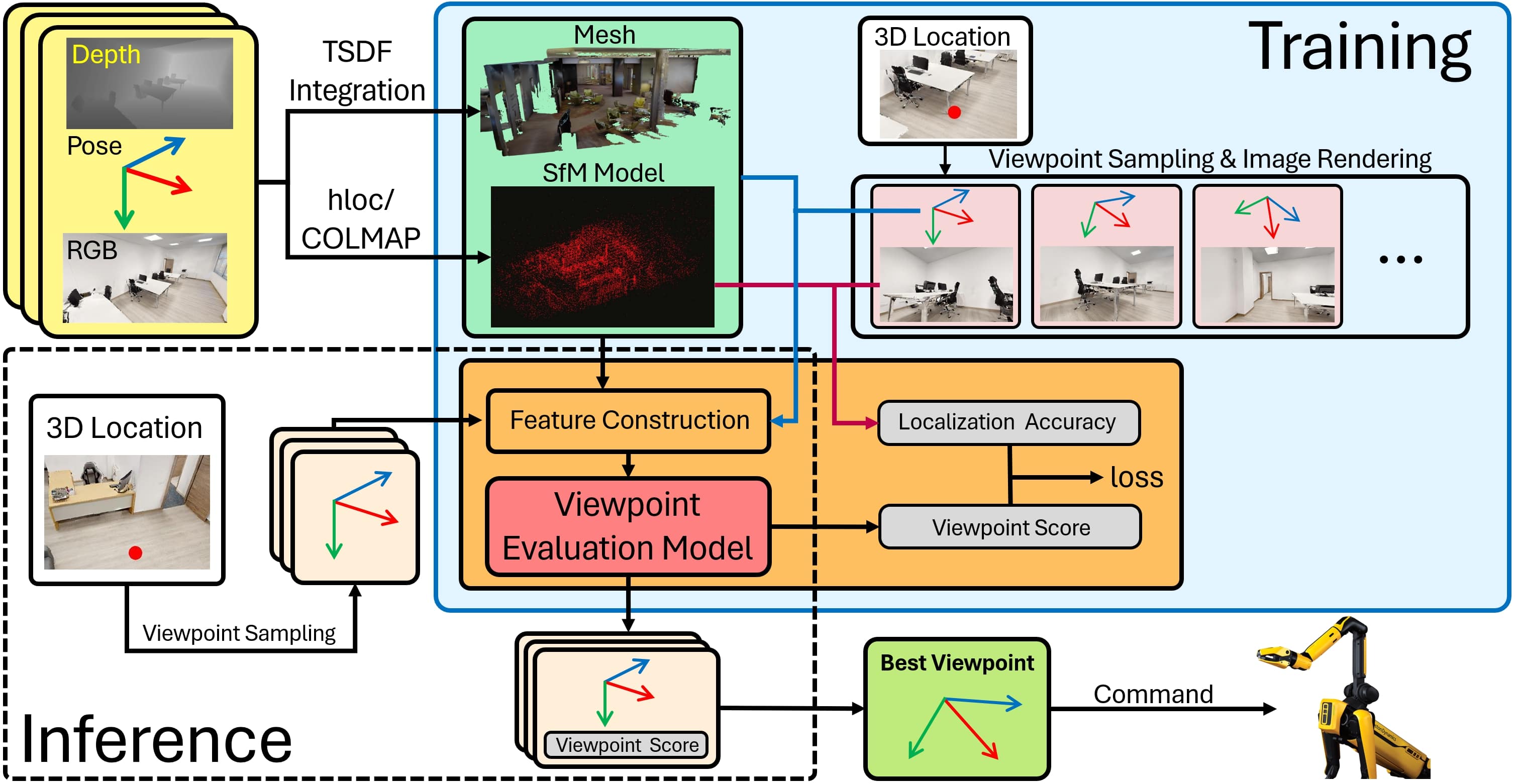}
    \caption{\textbf{Overview of the proposed active localization approach} The core of our approach is the learning-based viewpoint evaluation model. This model processes input features derived from an established Structure-from-Motion model alongside a camera viewpoint. It predicts the likelihood of the given viewpoint being effective for visual localization. In practice, when deployed, multiple viewpoints are sampled and assessed at a particular 3D location. The viewpoint that receives the highest predicted score is then chosen as the optimal one to execute for the robot.}
    \label{fig:diagram}
    \vspace{-5mm}
\end{figure*}

\textbf{Problem Statement}: Let $\bm{x} = (\bm{p},\bm{q})$, where $\bm{p} \in \mathcal{R}^3$ and $\bm{q} \in so(3)$ are the position and orientation of the robot sensor. Given a map representation of the environment $\mathcal{M}$ and a prior of the robot position $\hat{\bm{p}}$, our goal of the active visual localization is to find a viewpoint at that location, i.e., the orientation for the robot sensor, such that the visual localization method returns the most accurate estimation $\Bar{\bm{x}}$ at that location, i.e., 
\begin{align}
    \bm{q}^{\star} &= \argmin_{\bm{q}} \lVert \Bar{\bm{x}} - \bm{x} \rVert \label{eq:1} \\
   &= \argmin_{\bm{q}} \lVert \textit{loc} (\mathcal{M}, \mathcal{O}(\bm{x})) - \bm{x} \rVert \label{eq:2}
\end{align}
Here $\textit{loc}(\cdot)$ refers to different localization methods, which often take the observation $\mathcal{O}$, captured at the current pose $\bm{x}$, and localize it against the given map representation $\mathcal{M}$.  

In this paper, we construct $\mathcal{M}$ that combines the landmark point cloud $\mathcal{M}_{l}$ and the Truncated Signed Distance Function (TSDF) $\mathcal{M}_{t}$ of the environment. Any viewpoint selection policy $\uppi(\cdot)$ then takes $\mathcal{M} = (\mathcal{M}_{l},\mathcal{M}_{t})$ as prior knowledge of the environment, and selects a viewpoint for a certain estimated position:
\begin{align}
\label{eq:policy}
    \bm{q}^{\star} &= \uppi(\mathcal{M},\hat{\bm{p}})  
\end{align}
As an example, one of the baseline methods we implement takes the idea from~\cite{zhang2020fisher}:
\begin{align}
    \bm{q}^\textrm{FIM} &= \uppi_\textrm{FIM}(\mathcal{M},\hat{\bm{p}}) = \argmax_{\bm{q}}\sum \limits_{i \in \mathcal{M}_l} v(\bm{x}, i) \textrm{I}_i
\end{align}
where $v(\cdot)$ is the binary visibility of landmark $i$ and $\textrm{I}_i$ is the Fisher Information Metric (FIM) of observing landmark $i$. The visibility $v(\cdot)$ is determined using TSDF map $\mathcal{M}_{t}$. \\
% For a given viewpoint, a ray-casting step is performed, during which we filter out the landmarks that are in collision with the environment, with a small tolerance to account for inaccuracies in the process. 

We supplement this with two additional simple baselines:
\begin{align}
    \uppi_\textrm{max}(\mathcal{M},\hat{\bm{p}}) &= \argmax_{\bm{q}}\sum \limits_{i \in \mathcal{M}_l} v(\hat{\bm{x}}, i)\\
    \uppi_\textrm{angle}(\mathcal{M},\hat{\bm{p}}) &= \argmax_{\bm{q}}\sum \limits_{i \in \mathcal{M}_l} v_\textrm{angle}(\hat{\bm{x}}, i)
\end{align}
where $\uppi_\textrm{max}$ selects simply the view with the maximum visible landmarks, and $\uppi_\textrm{angle}$ uses a stricter visibility criterion inspired by~\cite{Davison2002SimultaneousVision} that only considers a landmark visible if its relative location to $\bm{x}$ is similar to those poses from which it was seen during mapping. 

% This is illustrated in Figure \ref{fig:angle_fig}.

We propose a data-driven approach, adhering to a "sample-and-evaluate" framework as depicted in Figure \ref{fig:diagram}. Our process involves gathering simulated data to train our viewpoint evaluation model, which is then tested through both simulated scenarios and real-world experiments. In the inference phase, we sample multiple viewpoints at a given location. For each viewpoint, we assemble its input feature vector by utilizing the landmark point cloud derived from the Structure-from-Motion (SfM) model. The rest of this section will detail our approaches, with the focus on constructing the feature vectors and the viewpoint evaluation model.
% \begin{figure}
%     \centering
%     \includegraphics[width=0.5\linewidth]{images/angle_criteria.pdf}
%     \caption{\textbf{Illustration of the Angle criteria}. Given \textcolor{blue}{poses} that see the landmark (red dot) during mapping, the \textcolor{green}{green pose} is within the directional range and has a higher chance of re-detecting the landmark. The \textcolor{red}{red pose} however sees the landmark from a new direction and does not count it based on the criteria.}
%     \label{fig:angle_fig}
%     \vspace{-4mm}
% \end{figure}

\textbf{Data-driven viewpoint evaluation} For each viewpoint, we collect the following essential information:
\begin{itemize}
    \item Its distance to every landmark of the map.
    \item The viewing angle between every landmark and the principle axis. 
    \item The minimum and maximum distance and viewing angle per landmark has, with respect to the camera frame during the mapping stage
    \item Pixel coordinates of every landmark in the camera frustum of $\bm{x}$
    \item The number of landmarks in the previously seen angle range
    \item Its corresponding DINO~\cite{oquab2023dinov2} appearance features for every landmark
\end{itemize}
 To collect training data, we generate for every viewpoint the above information, as well as the ground truth pose and the result of the visual localization method at that viewpoint. Based on this data, we train our model to classify whether a viewpoint has been localized within an error threshold. In particular, we distinguish between two methods, one is based on Multi-Layer Perceptron (MLP) $\uppi_{MLP}$, and the other one is based on the Transformer $\uppi_{VPT}$. The way of encoding the collected information differs respectively. We design lightweight models in order to maintain online capability.  Both kinds of models end with a softmax layer to determine the viewpoint score for classification.  

The model based on  MLP requires a fixed input dimensionality. However, the number of landmarks varies across different viewpoints, we set a feature aggregation step in the model of $\uppi_{MLP}$. For each information in the list above, we build a histogram to aggregate the corresponding value of all the filtered landmarks. For the pixel coordinates, we aggregate the information in a 2D heatmap instead of a 1D histogram. DINO features cannot be processed in the MLP-based model.

The constraints of fixed input dimensionality also encourages our motivation to investigate the transformer architecture. With the transformer, our inputs can be provided in a per-landmark fashion, which also allows for the inclusion of features that cannot be easily aggregated, such as DINO appearance per landmark.

% \textbf{Viewpoint Evaluation}
Upon obtaining the trained model, for a specified point $\bm{p}$, we assess all the sampled viewpoints. Each candidate is allocated a localization score, reflecting its efficacy in visual localization. Subsequently, we select the viewpoint that is awarded the highest score. For instance:

\begin{align}
   \uppi_\textrm{VPT}(\mathcal{M},\hat{\bm{p}}) &= \argmax_{\bm{q}} f_{VPT} (\mathbf{F}(\hat{\bm{x}}, \mathcal{M}))
\end{align}
where $\mathbf{F}$ are the input features, and $f_{VPT}(\cdot)$ returns the output score from the viewpoint transformer (VPT) model.
% For each specific approach(MLP-based or VPT-based), its scoring function $\text{ SCORE}_{(\cdot)}(\cdot)$ takes the corresponding feature inputs ($\mathbf{F}_{MLP}$ or $\mathbf{F}_{VPT}$), gathered at a certain position and viewpoint $\mathbf{x}$, and predicts the score that indicates how well its localization approach will perform. The model architecture of the score functions for both approaches can be found in Fig XXX. 

% \begin{figure}
%     \centering
%     \includegraphics[width=1.0\linewidth]{images/Linear_Model.pdf}
%     \includegraphics[width=1.0\linewidth]{images/PCT_Model-1.pdf}
%     \caption{\textbf{Model architecture.} The MLP model(upper) consists of two 1D-convolutions, followed by a sequence of seven fully connected linear layers with progressively smaller output sizes. The VPT model(lower) starts with an initial linear embedding, The encoded features are then passed into two Self-Attention heads and the outputs of each are concatenated. The resulting feature map is then passed to a sequence of linear layers. Both models have a final stage that outputs two logits and feeds them into the SoftMax function to calculate the viewpoint score. In total, the MLP model has approximately 181k trainable parameters. The VPT model has approximately 159k parameters.}
%     \label{fig:MLP_model}
% \end{figure}

\section{Experiments}
We develop a comprehensive pipeline for data collection aimed at training, validation, and testing across simulated and real-world settings, and we carry out a variety of experiments. Additionally, we implement several baseline methods for comparison. The data collection and model training is mostly done within the simulation, details can be found in \ref{Exp:data} and \ref{Exp:training}. We present all baseline methods alongside our evaluation strategy in \ref{Exp:evl}. Finally, our real-world experiment findings are detailed in \ref{Exp:realworld}.

\subsection{Data Generation}
\label{Exp:data}
\textbf{Simulated Scenario} We focus on indoor scenarios, and choose a selection of scenes from the Habitat-Matterport 3D (HM3D) dataset~\cite{ramakrishnan2021habitat}. HM3D provides high-quality 3D reconstructions of real-world indoor environments with textured 3D meshes. We selected nine scenes, as depicted in Figure~\ref{fig:env_overview}, and imported them into NVIDIA Isaac Sim. Subsequently, we manually captured controlled trajectories using a simulated camera setup that mimics the Microsoft HoloLens 2~\cite{hubner2020evaluation}, ensuring the camera was positioned at a height akin to that of a human.
%Specifically, nine scenes are chosen from HM3D, based on the fact that they contained only a single floor and represented a commercial space, such as a store or office, neglecting scenes from private residencies due to the abundance of tight spaces and stairs, which would make environment traversal unnecessarily complex for the case studied in this work. An overview of the chosen environments can be seen in Figure XXX. We imported these scenes into NVIDIA Isaac Sim in order to create configurable sensors and let them interact with the scenes. For the specific sensor configuration we use, in order to collect data that closely resemble those that could be generated within a real-world human-robot collaboration situation, two different camera rigs are designed. One is replicated using intrinsic and extrinsic values from the Microsoft HoloLens 2. The sensor rig specification of HoloLens 2 can be found in figure \ref{fig:hl-sensors}. Following its configuration, this simulated sensor rig incorporates the four forward and side-facing cameras, with a resolution of 480x640, and the intrinsic parameters taken from the calibration process of the device. The second sensor rig contains a single RGB-D camera, which serves as an onboard sensor of the robot.   

% \begin{figure}
%     \includegraphics[width=1.0\linewidth]{images/HL_sensors.pdf}
%     \caption{\textbf{Overview of available HoloLens 2 sensors} \cite{Hubner2020EvaluationApplications}}
%     \label{fig:hl-sensors}
% \end{figure}

\textbf{Data Collection} To acquire $\mathcal{M}_{t}$, we utilize the depth images captured by sensors and conduct TSDF (Truncated Signed Distance Field) integration\footnote{Isaac Sim offers a direct method to ascertain the occupancy details of a 3D scene through its built-in functionality, which we also employ as an alternative approach.}. Both the generated meshes and the occupancy maps can be used for  occlusion and collision checking.

To obtain $\mathcal{M}_{l}$, we use mapping and localization frameworks from hloc and COLMAP~\cite{sarlin2019coarse,Schonberger2016Structure-from-MotionRevisited} to extract 2D local features from images and build a 3D landmark point cloud. The collected images from the simulated HoloLens 2 are fed into the pipeline to build the 3D landmarks map. The per-landmark feature vector $\mathbf{F}$ are also created and attached to each landmark at this stage. 

% Additionally, landmarks that do not fit the following criteria are removed with the aim of improving map quality:
% \begin{itemize}
%     \item observed in at least 4 mapping images
%     \item observation distance within the range [0.1m, 8.0m]
% \end{itemize}
 For running visual localization, We mainly use the localization module from hloc as the framework for $ \textit{loc}(\cdot)$ in~\ref{eq:2}.

To recover the scale of the map, one option would be to create the reconstruction with known camera poses. However, this would not be representative of how this process could be done using real hardware, where poses would be estimated, for example, from a visual SLAM module. In our experiment, all maps are created without the exact camera poses. Instead, we build the reconstructions from the images alone and then using RANSAC to estimate the scale with respect to the ground-truth poses.    

To collect training data point in the simulator, we create random camera paths at a height characteristic of robots, using a single RGB-D camera to mimic an onboard robot sensor. At each waypoint, we capture a set of viewpoint samples and store the images along with their exact viewpoint poses as ground truth. As illustrated in~\ref{eq:1} and \ref{eq:2}, we then employ the localization module for each sample, calculating the discrepancy between the estimated pose and the actual ground truth pose.

%%%% SUPERPOINT + SIFT TABLE %%%
\begin{table*}
\centering
\resizebox{1.0\textwidth}{!}{%
\begin{tabular}{cl|rrrrrr}
\toprule
  \multicolumn{2}{l|}{distance [m]} & 0.025 & 0.05 & 0.075 & 0.1 & 0.25 & 1.0\\
  \multicolumn{2}{l|}{orientation [deg]} & 1 & 1 & 1 & 1 & 2 & 5\\
  \midrule
\multirow{8}{*}{\rotatebox{90}{w/o occl. filt.}} & Forwards & 62.57                      & 77.25 & 80.84 & 81.99 & 85.23 & 86.68 \\ 
& Random & 54.84                      & 69.31 & 72.60 & 73.90 & 78.29 & 80.29 \\ 
& max & 60.43                      & 76.05 & 79.29 & 80.64 & 82.93 & 84.98 \\
& angle & 74.35                      & 85.58 & 87.28 & 87.72 & 89.07 & 90.27 \\ 
& MLP & 76.25                      & 86.78 & 88.37 & 88.82 & 90.42 & 91.52 \\ 
& VPT & 74.75                      & 87.67 & 90.67 & 91.27 & 92.96 & 93.86 \\ 
& VPT + DINO & 69.16                      & 84.68 & 87.97 & 88.82 & 90.92 & 92.07 \\ \midrule
\multirow{5}{*}{\rotatebox{90}{w/ occl. filt.}} & max & 70.21                      & 84.98 & 88.02 & 89.42 & 91.37 & 92.22 \\ 
& angle & 78.34                      & 88.22 & 90.02 & 90.37 & 91.52 & 92.56 \\ 
& FIM & 65.17                      & 78.54 & 81.79 & 83.53 & 86.83 & 88.67 \\
& MLP & \textcolor{red}{79.69}                      & 89.72 & 91.42 & 92.17 & 93.66 & 94.91 \\ 
& VPT& \textcolor{blue}{79.54}                      & \textcolor{red}{90.97} & \textcolor{red}{93.31} & \textcolor{red}{93.76} & \textcolor{red}{94.61} & \textcolor{blue}{95.21} \\ 
& VPT + DINO  & 78.64                      & \textcolor{blue}{90.52} & \textcolor{blue}{92.56} & \textcolor{blue}{92.91} & \textcolor{blue}{94.56} & \textcolor{red}{95.36} \\ \midrule
& Best Possible & 96.31                      & 97.46 & 97.70 & 97.70 & 98.05 & 98.10 \\ \bottomrule
\end{tabular}%
\quad
\begin{tabular}{cl|rrrrrr}
\toprule
  \multicolumn{2}{l|}{distance [m]} & 0.025 & 0.05 & 0.075 & 0.1 & 0.25 & 1.0\\
  \multicolumn{2}{l|}{orientation [deg]} & 1 & 1 & 1 & 1 & 2 & 5\\
  \midrule
\multirow{8}{*}{\rotatebox{90}{w/o occl. filt.}} & Forwards & 37.97                      & 55.96 & 62.62 & 65.71 & 74.70 & 79.22 \\ 
& Random & 31.21                      & 48.66 & 55.57 & 57.80 & 65.46 & 70.23 \\ 
& max & 33.05                      & 50.94 & 57.60 & 60.19 & 67.99 & 73.01 \\ 
& angle & 41.20                      & 55.86 & 60.59 & 63.07 & 70.73 & 75.60 \\ 
& MLP & \textcolor{blue}{51.09}                      & 67.84 & 73.56 & 75.84 & 81.61 & 84.69 \\ 
& VPT & 44.23                      & 62.77 & 69.48 & 72.12 & 78.98 & 82.85 \\ 
& VPT + DINO & 39.02                      & 58.45 & 65.16 & 68.34 & 76.44 & 80.27 \\ \midrule
 \multirow{5}{*}{\rotatebox{90}{w/ occl. filt.}} & max & 46.42                      & 67.84 & 74.20 & 76.54 & 84.05 & \textcolor{blue}{88.72} \\ 
& angle & 50.80                      & \textcolor{red}{70.78} & 75.75 & 78.03 & 84.59 & 88.52 \\
& FIM & 43.34                      & 63.02 & 69.53 & 70.87 & 78.43 & 82.85 \\ 
& MLP & \textcolor{red}{51.49}                      & \textcolor{blue}{70.43} & \textcolor{red}{76.94} &\textcolor{red}{ 79.47} & \textcolor{blue}{84.94} & 88.02 \\ 
& VPT & 41.70                      & 64.12 & 71.67 & 74.75 & 81.76 & 85.74 \\ 
& VPT + DINO & 47.27                      & 69.48 & \textcolor{blue}{76.49} & \textcolor{blue}{79.37} & \textcolor{red}{85.14} &\textcolor{red}{ 89.26} \\ \midrule
& Best Possible & 88.52                      & 92.84 & 93.74 & 94.28 & 97.02 & 97.66 \\ \bottomrule
\end{tabular}%
}

\caption{\textbf{Evaluation of viewpoint strategies} in the simulated testing environments using both \textbf{SuperPoint}(Left) and \textbf{SIFT}(Right) features.  Recall percentages are shown at varying distance and orientation thresholds, highlighting the \textcolor{red}{best} and \textcolor{blue}{second-best} performing methods.  Rows marked as `with occlusion filter' are those where landmarks are filtered with the occlusion. The bottom row is an oracle method that selects the viewpoint with smallest possible error among all samples.}
\label{tab:results}
\vspace{-4mm}
\end{table*}

\subsection{Model Training} 
\begin{figure}
  \centering
  \footnotesize
  \includegraphics[width=0.96\linewidth]{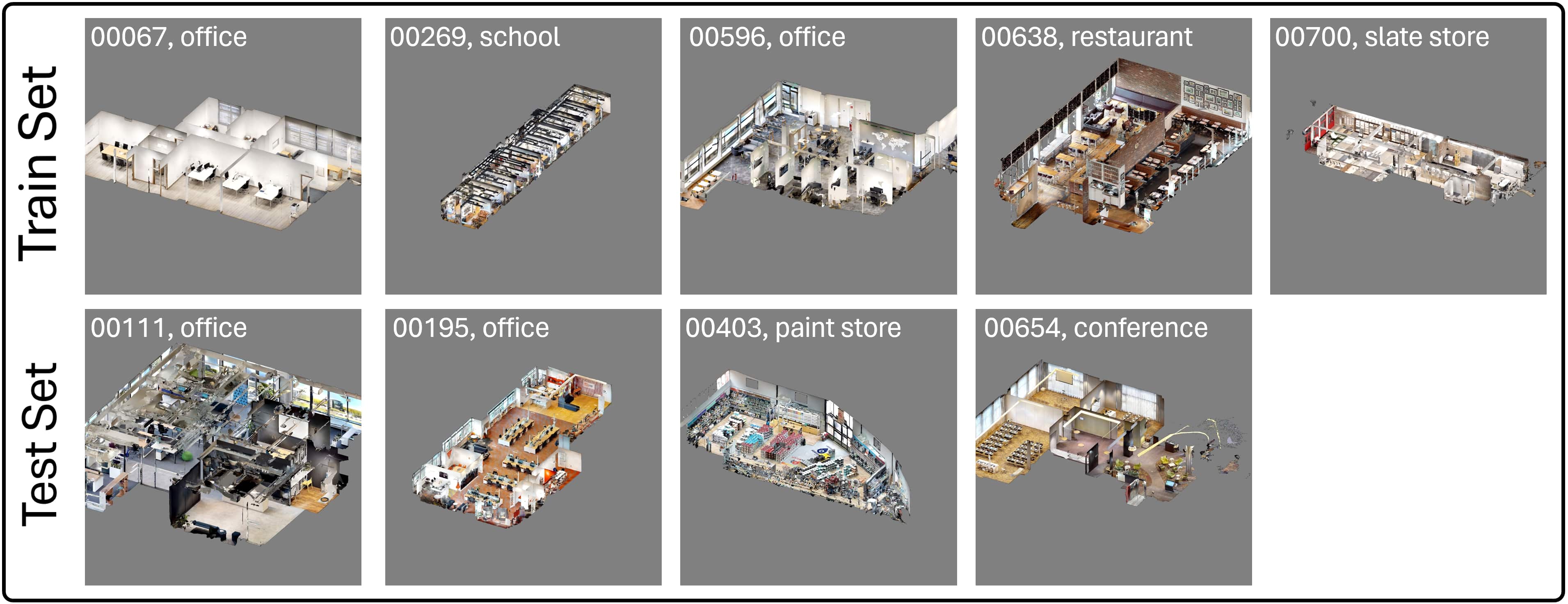}
  \caption{\textbf{Overview of the dataset.} The number in brackets following the designation corresponds to the index in the Habitat-Matterport 3D dataset. A small collection of scenes lead to great generalization capability of our model, thanks to our effective data point sampling method.}
  \label{fig:env_overview}
  \vspace{-1mm}
\end{figure}

We train both two kinds of models on the classification task of predicting whether a viewpoint will result in localization errors smaller than 0.1m and 1 deg. The primary preprocessing actions include normalizing all input features $\mathbf{F}$ to a range between 0 and 1 and ensuring the dataset has a balanced distribution of positive and negative examples. The division of the training and validation datasets is illustrated in Figure~\ref{fig:env_overview}. For every scene, we randomly generate 100 waypoints. At each waypoint, we sample 50 viewpoints, allowing complete rotation around the yaw axis while keeping the pitch angle within a degree range between -10 to 45. This procedure generates a training and validation dataset comprising 25,000 images and a test set containing 20,000 images.

\label{Exp:training}
\subsection{Evaluation Strategy}
\label{Exp:evl}

As detailed in \ref{sec:method}, we implement three baseline strategies $\uppi_\textrm{FIM}, \uppi_\textrm{max}, \uppi_\textrm{angle}$, and augment them with two additional simplistic approaches: $\uppi_\textrm{forward}$, which directs the camera towards the next waypoint in the trajectory, and $\uppi_\textrm{random}$, which chooses viewpoints at random. Following the methodology established during the training phase, random waypoints and viewpoint samples are generated for each evaluation setting. Each sampled viewpoint is evaluated according to the different viewpoint-selection strategies, recording the localization error for the selected viewpoints. To establish an upper limit for the attainable localization precision at a given position, the minimal error observed from the localization for each sample is also recorded and treated as the pseudo-optimal viewpoint selection strategy.

\begin{figure}
    \centering
    \includegraphics[width=0.85\linewidth]{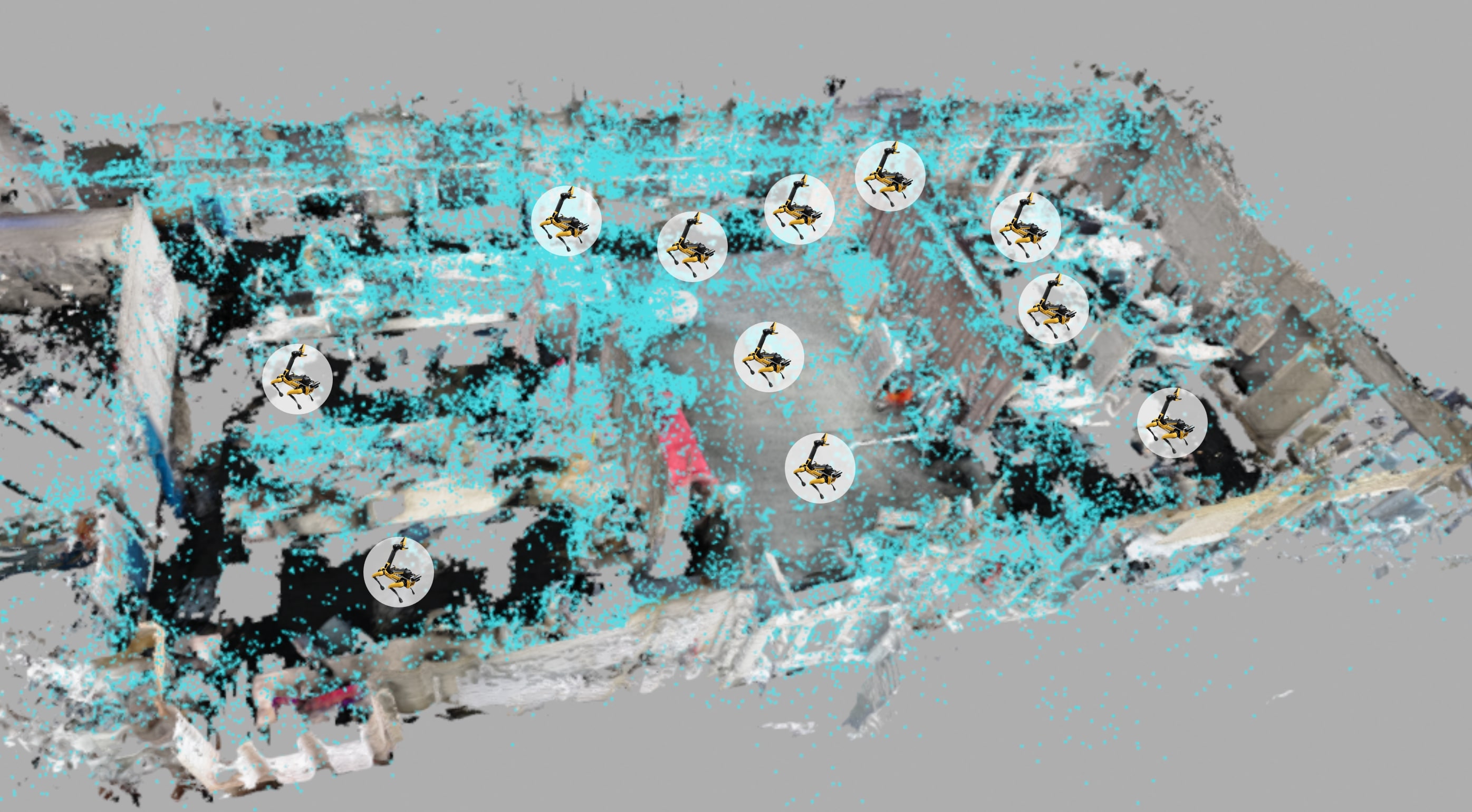}
    \caption{\textbf{The constructed map for real-world evaluation} The landmarks point cloud (blue) is aligned with the environment mesh. Evaluated locations are shown as mini robots.}
    \label{fig:spot_eval_locations}
    \vspace{-5mm}
\end{figure}

\subsection{Real-world Deployment and Test}
\label{Exp:realworld}
\textbf{Setup} Tthe best-performing methods are implemented into a ROS-compatible planning module and deployed on a quadruped robot with a robotic arm that contains a calibrated color camera in its end effector, allowing different viewpoints to be viewed for a given body position. We use a HoloLens 2 to build a map of an indoor environment. We use a ROS-compatible hloc implementation from \cite{Suomela_2023_WACV}.

\textbf{Data Collection} In contrast to the simulated environment where ground-truth poses are readily available, real-world deployments lack these precise positional references. To assess the effectiveness of various viewpoint-choosing strategies in such settings, two kinds of positional data are essential: Each viewpoint planner must first obtain an initial estimate $\hat{\bm{p}}$, which is utilized in~\ref{eq:policy} to determine an appropriate viewing direction at the current location. Following the selection of a viewing direction, the robot adjusts its arm to capture an image, which is then localized in relation to the pre-existing map. Subsequently, a ground-truth pose is necessary to validate the accuracy of this localization

To measure accurate poses in the map created from the HoloLens recording, we use a combination of visual localization and AprilTag fiducial markers~\cite{olson2011apriltag}. Once a map of the environment has been created, an AprilTag is affixed to a static location. We capture several high-quality images of the marker, including its surroundings, using a calibrated DSLR camera. The AprilTag marker can be accurately localized within the captured images, and by ensuring that the images contain enough of the surroundings, they can also be localized in the map using visual localization. This two-step process provides an estimated location of the AprilTag marker within the map for each image. An average of the estimated locations is taken in order to minimize the effect of errors in the tag or localization. This process gives us accurate poses of fiducial markers with respect to the captured landmark map. 

For any waypoint on which we want to evaluate the viewpoint strategies, we initialize the robot at the location where we placed a fiducial marker and walk the robot to the investigated waypoint. We estimate $\hat{\bm{p}}$ of that waypoint from observing the fiducial marker with the robot camera and integrating the odometry to the investigated waypoint. This resembles a realistic odometry-based localization prior that can be fed into the viewpoint-choosing strategies.

% A similar approach can then be used to obtain the best-possible estimate of the robot body pose at any position in the map by placing a marker on a stable tripod that can be detected in both the robot's end-effector camera and in a collection of calibrated DSLR images. We use the body pose of the robot estimated in that way, together with the inverse kinematics of the arm, as the ground truth against which we compare the outcomes of visual localization.

% Overall, we use the following process for evaluating the viewpoint-choosing strategies in a real-world environment:
% \begin{enumerate}
%     \item get initial robot body pose from static localization tag
%     \item move the robot to the measurement location
%     \item move end-effector camera on the robot amr to viewpoints from the evaluated strategies and record images, as well as arm inverse kinematics
%     \item place a marker in front of the robot and localize it with respect to the robot camera and with respect to the map using DSLR images
%     \item estimate robot body positions based on recorded images and compare it to best-possible localization
% \end{enumerate}

\section{Results}

\begin{figure}
    \centering
    \vspace{+1mm}
    \includegraphics[width=0.84\linewidth]{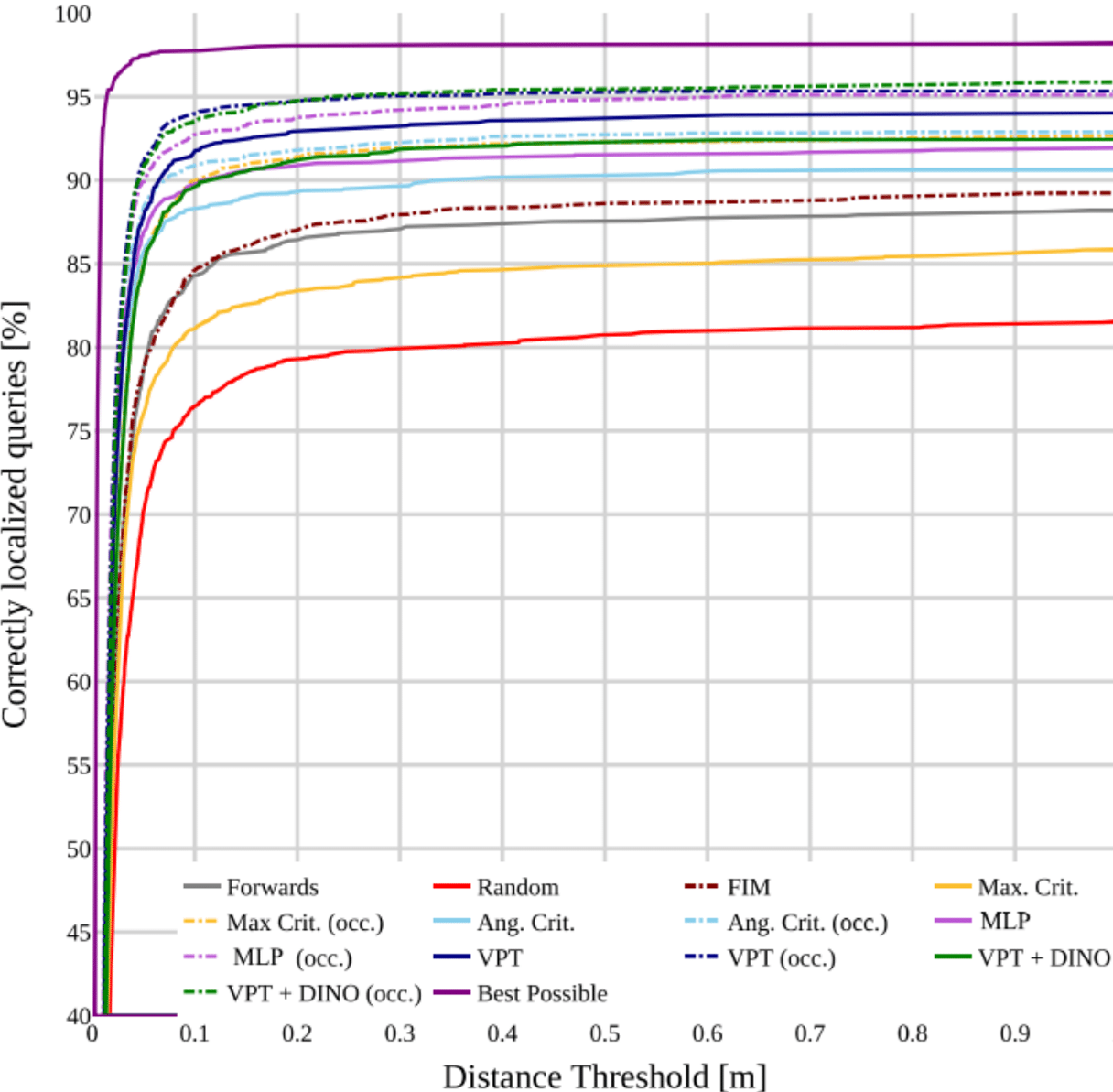}
    \caption{\textbf{Cumulative distribution of position errors} of the evaluation points in the testing environments using \textbf{SuperPoint} feature. Methods or versions of methods that rely on the environment mesh in order to determine landmark occlusion are plotted using \textbf{$\cdot-\cdot-$} lines.}
    \label{fig:SP_CFD}
    \vspace{-6mm}
\end{figure}

We evaluate our approaches and the baseline methods in both our simulation pipeline and the real world. Table \ref{tab:results} shows the quantitative results from the simulation. To demonstrate the generalization ability of our approaches across different feature descriptors, we evaluate all methods using both SuperPoint feature~\cite{detone2018superpoint} and SIFT feature~\cite{Lowe1999ObjectFeatures}, whereas both of our scoring models are only trained with the SuperPoint feature. The result in the table shows that, with SuperPoint feature, our methods perform the best under every error level, even though our models are only trained to classify the 0.1 m, 1 deg error level. When we switch from SuperPoint to SIFT, recall percentages drop for all the methods, however, our data-driven methods still have the highest recall at almost every error level.

To explore the impact of DINO features and occlusion handling on model performance, we conduct experiments with various model configurations, including MLP-based and VPT-based models that do not pre-filter based on occlusion information, as well as VPT-based models that do not incorporate DINO features in their input. The findings reveal that neglecting occlusion leads to diminished performance for both model types, a trend that is also observed in baseline methods. Interestingly, the VPT-based model demonstrates the ability to exceed the efficacy of traditional non-data-driven approaches at its designated training threshold, even without occlusion considerations. We accumulate the correctly-localized waypoints along with different distance thresholds and show the result in Figure~\ref{fig:SP_CFD}, where it is easier to verify that our learning-based approaches outperform the other baseline methods with both SuperPoint and SIFT features.

% It is also easy to observe the performance improvement by considering the occlusion.

Results from the 12 waypoints (see Figure~\ref{fig:spot_eval_locations}) in our real-world experiment are shown in Table~\ref{tab:Spot_results}. Although the sample size is relatively small, we still see the dominating performance of the VPT-based model, which shows its generalization ability to the real world. Besides, our final approach shows great online performance. It achieves to successfully select viewpoints from 100 candidates in less than one second, running on a regular workstation with a NVIDIA GeForce RTX 2080 GPU. 

The evaluation results highlight the effectiveness of data-driven methods in comparison to hand-crafted information metrics. Learning decision boundaries for landmark features and encoding diverse information enable data-driven approaches to outperform traditional heuristic metrics. Despite being trained with a limited dataset, these scoring models exhibit robustness across various scenarios, different local features, and both simulated and real-world environments. This underscores the potential of data-driven approaches, which adopt a less heuristic approach. 

We note variations in performance between the two machine learning methodologies. The transformer model exhibits reduced generalization across distinct feature descriptors yet demonstrates superior transferability from simulated data to real-world applications. Conversely, the MLP-based model and the VLP-based model employing DINO underperform with SuperPoint features and in real-world evaluations but show enhanced adaptability to SIFT features. This suggests that in these scenarios, they benefit from a more substantial semantic prior provided by DINO, which in turn offers greater generality across various feature types.

In conclusion, our proposed data-driven light transformer-based model exhibits optimal performance when evaluated on the same feature descriptors as those used during training. The inclusion of occlusion filtering based on mesh reconstruction enhances the performance of all approaches. It's important to note that in cases where geometric data isn't available, the proposed data-driven approach still produces superior results. This could be attributed to the learning process capturing information about the angle ranges from which landmarks were observed, which indirectly encodes geometric information.

\begin{table}
\vspace{2mm}
\centering
% \resizebox{0.5\textwidth}{!}{%
\begin{tabular}{l|rrr}
  \toprule
  distance [m] & 0.25 & 0.5 & 1.0\\
  orientation [deg] & 2.0 & 3.5 & 5.0\\
  \midrule
  Arm stowed & 0.0   & 25.0 & 41.67 \\
Forwards & \textcolor{blue}{16.67}   & 50.0 & 58.33 \\ 
Robot odometry & 0.0   & 33.33 & 66.67 \\ 
Random & 8.33   & 41.67 & 58.33 \\ 
FIM & 8.33   & 50.0 & 75.0 \\ 
max & 8.33   & \textcolor{blue}{66.67} & \textcolor{blue}{83.33} \\ 
angle & 8.33   & 58.33 & 66.67 \\ 
MLP & 0.0   & 58.33 & 58.33 \\ 
VPT & \textcolor{red}{33.33}   & \textcolor{red}{83.33} & \textcolor{red}{91.67} \\ 
\bottomrule
\end{tabular}%
% }
\caption{\textbf{Evaluation of viewpoint strategies} in the real-world environment with the quadruped robot. The table layout is the same as Table \ref{tab:results}}
\label{tab:Spot_results}
\vspace{-2mm}
\end{table}

\section{Conclusion}
This work addresses the challenge of localizing ground robots within an existing map constructed from devices with varying perspectives.We introduce a novel data-driven approach to explore effective utility functions for this task and evaluate it alongside diverse viewpoint selection methods in the literature. Experiment results show that our approach greatly improves robot localization ability within a known point-cloud map in the presence of large viewpoint changes and occlusion from ground-level obstacles. These improvements are observed in both simulated and real-world environments.

% Future endeavors could focus on further enhancing the generalization of our approach across diverse localization methods, map representations, and environmental conditions. Additionally, there is potential for addressing more complex situations, such as dynamic scenarios, to broaden the scope of applicability.

\newpage

\bibliographystyle{IEEEtran}
\bibliography{references}

\end{document}